\documentclass[conference]{IEEEtran} \IEEEoverridecommandlockouts
\usepackage{cite}
\usepackage{amsmath,amssymb,amsfonts}
\usepackage{algorithmic}
\usepackage{booktabs}
 \usepackage[export]{adjustbox}
\usepackage{graphicx}
\usepackage{textcomp}
\usepackage{xcolor}

\usepackage{fancyhdr}

\fancypagestyle{icecsheader}{
    \fancyhf{}
    \fancyhead[C]{%
        \fontsize{8}{9}\selectfont
        2026 IEEE 33rd International Conference on Electronics, Circuits and Systems (ICECS)}

}

\fancypagestyle{icecsfirstpage}{
    \fancyhf{}
    \fancyhead[C]{%
        \fontsize{8}{9}\selectfont
        2026 IEEE 33rd International Conference on Electronics, Circuits and Systems (ICECS)}
    \fancyfoot[L]{%
        \fontsize{8}{9}\selectfont
        979-8-3195-1905-4/26/\$31.00 \textcopyright{}2026 IEEE}

}

\renewcommand{\baselinestretch}{0.875}
\def\BibTeX{{\rm B\kern-.05em{\sc i\kern-.025em b}\kern-.08em
    T\kern-.1667em\lower.7ex\hbox{E}\kern-.125emX}}
\begin{document}

\title{Co-Optimizing Graph Sparsification and Approximate Computing for Energy-Efficient FPGA-Based GCN Inference
}

\author{
    \IEEEauthorblockN{Nathaniel Kaye Mellor, Shreejith Shanker, George Floros
    \IEEEauthorblockA{
    Department of Electronic and Electrical Engineering, Trinity College Dublin, Ireland\\
\{kayemeln, shankers, florosg\}@tcd.ie
}}}

\maketitle

\begin{abstract}
Graph Convolutional Networks (GCNs) have emerged as a powerful framework for learning from graph-structured data, yet their deployment on resource-constrained edge platforms remains challenging due to the computational and memory demands of sparse graph aggregation. This work presents an FPGA-based GCN accelerator that combines DSpar graph sparsification, 8-bit quantization, and approximate multipliers on the AMD Kria KV260. Evaluated on Cora, LastFM Asia, and Amazon Photo, the design explores the interaction between sparsification and approximation across graphs with widely varying densities.
Results show that the effectiveness of approximate arithmetic is governed by accumulation depth within GCN computations. Approximate multipliers are most effective when applied to sparse aggregation operations, while graph sparsification further improves their viability by reducing aggregation depth. The combined approach achieves up to 9.88$\times$ speedup while maintaining 86.6\% classification accuracy on Amazon Photo, and 1.52$\times$ speedup with 77.0\% accuracy on Cora, with total power consumption below 1 W. These results demonstrate that graph sparsification and approximate computing are complementary techniques whose co-optimization enables efficient low-power GCN inference on edge FPGA platforms.
\end{abstract}


\section{Introduction}
Graph Neural Networks (GNNs) have emerged as a powerful class of machine learning models for learning from graph-structured data, achieving state-of-the-art performance across applications including recommender systems, robotics, traffic forecasting, molecular analysis, and anomaly detection \cite{corso2024graph}. Among the various GNN architectures, Graph Convolutional Networks (GCNs) have gained widespread adoption due to their simplicity and strong predictive performance \cite{wu2019simplifying}. As graph-based learning continues to move into latency-sensitive and privacy-critical domains, efficient deployment of GCN inference on edge devices has become increasingly important.

However, deploying GCNs on resource-constrained edge platforms remains challenging \cite{zeng2025edge}. A typical GCN layer consists of a sparse aggregation operation followed by a dense feature transformation. The aggregation stage is commonly implemented as a sparse-dense matrix multiplication (SpMM) involving the graph adjacency matrix, which is inherently sparse and characterized by irregular memory access patterns and highly skewed node-degree distributions \cite{wu2022skeletongcn}. As a result, sparse aggregation often dominates inference latency and energy consumption, limiting the efficiency of conventional processing platforms.

FPGAs have emerged as an attractive platform for GCN acceleration due to their energy efficiency and reconfigurability. Recent FPGA-based GCN accelerators address the challenges of sparse and irregular graph computation through a variety of architectural techniques. LW-GCN employs a software-hardware co-design approach with custom sparse formats, workload balancing, and quantization \cite{tao2022lw}, while BoostGCN improves memory efficiency through graph partitioning and pipelined execution \cite{zhang2021boostgcn}. Other works exploit graph properties such as adjacency matrix symmetry for improved data reuse \cite{nair2023fpga} or graph structural similarity to reduce convolution operations \cite{romero2021ace}. While these approaches improve throughput and energy efficiency, they primarily focus on optimizing sparse computation and memory access patterns. The combined use of graph sparsification and approximate arithmetic remains largely unexplored.

The contributions of this work are as follows. \textit{First}, we present an FPGA-based GCN accelerator that combines DSpar graph sparsification \cite{liu2023dspar}, 8-bit quantization, and approximate multipliers within a unified HLS framework. To the best of our knowledge, this is the first integration of DSpar and approximate multipliers within a dedicated GNN accelerator. \textit{Second}, we show that the effectiveness of approximate multipliers is governed by accumulation depth, and that graph sparsification improves their viability by reducing aggregation depth. \textit{Third}, we demonstrate this interaction across three datasets spanning a wide range of graph densities, from Cora (4.9 average neighbors per node) and LastFM Asia (8.3) to Amazon Photo (32.1), showing that the benefits of combining sparsification and approximation become increasingly pronounced on denser graphs. \textit{Finally}, the proposed design achieves up to 9.88$\times$ speedup while maintaining 86.6\% classification accuracy on Amazon Photo, with total power consumption below 1W.



The remainder of this paper is organized as follows. Section \ref{sec:back} introduces the GCN inference model and the graph sparsification technique used in this work. Section \ref{sec:proposed} presents the proposed accelerator architecture and implementation methodology. Section \ref{sec:Exp} describes the experimental setup and evaluation methodology. Finally, Section \ref{sec:Concl} concludes the paper.

\begin{figure*}[!hbt]
    \begin{center}
        \includegraphics[width=1.95\columnwidth]{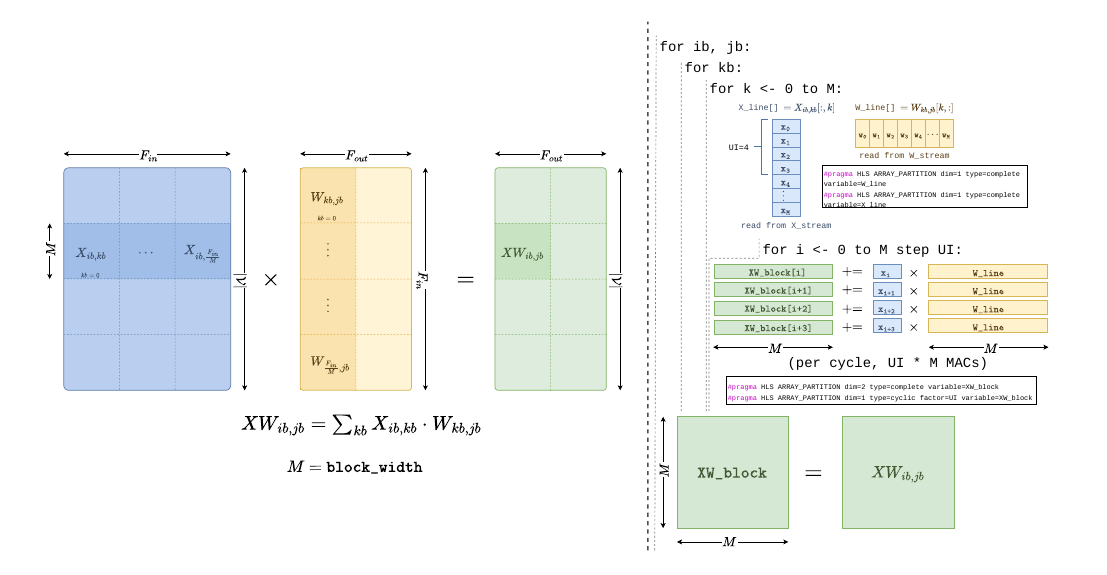}
        \caption{Tiled GEMM architecture showing data reuse, loop unrolling, and parallel MAC execution for feature transformation.}
\label{fig:gemm_figure}
    \end{center}
\end{figure*}
\section{GCN Inference Challenges} \label{sec:back}

GCNs are a widely used class of GNNs that learn node representations by aggregating information from neighboring vertices. Let $G=(V,E)$ denote a graph with adjacency matrix $\mathbf{A}$ and node feature matrix $\mathbf{X}$. The forward propagation of the $l$-th GCN layer is defined as

\begin{equation}
\mathbf{H}^{(l+1)}
=
\mathrm{ReLU}
\left(
\tilde{\mathbf{A}}
\mathbf{H}^{(l)}
\mathbf{W}^{(l)}
\right),
\label{eq:gcn_layer}
\end{equation}

where $\mathbf{H}^{(l)}$ denotes the node embedding matrix at layer $l$, with $\mathbf{H}^{(0)}=\mathbf{X}$, and $\mathbf{W}^{(l)}$ is the trainable weight matrix. The normalized adjacency matrix is defined as

\begin{equation}
\tilde{\mathbf{A}}
=
\tilde{\mathbf{D}}^{-\frac{1}{2}}
(\mathbf{A}+\mathbf{I})
\tilde{\mathbf{D}}^{-\frac{1}{2}},
\label{eq:norm_adj}
\end{equation}

where $\mathbf{I}$ is the identity matrix and $\tilde{\mathbf{D}}$ is the degree matrix of $(\mathbf{A}+\mathbf{I})$. For hardware implementation, (\ref{eq:gcn_layer}) is commonly decomposed into a dense matrix multiplication (GEMM)

\begin{equation}
\mathbf{M}
=
\mathbf{H}^{(l)}
\mathbf{W}^{(l)},
\label{eq:gemm}
\end{equation}
followed by a sparse-dense matrix multiplication (SpMM)

\begin{equation}
\mathbf{H}^{(l+1)}
=
\mathrm{ReLU}
\left(
\tilde{\mathbf{A}}
\mathbf{M}
\right),
\label{eq:spmm}
\end{equation}
where the GEMM operation performs feature transformation and the SpMM operation performs neighborhood aggregation. Since the computational complexity of SpMM scales with the number of non-zero entries in $\tilde{\mathbf{A}}$, sparse aggregation often dominates the execution time of GCN inference.

To reduce the computational cost of graph aggregation, we employ DSpar \cite{liu2023dspar}, a lightweight graph sparsification technique that removes edges prior to inference. Unlike traditional spectral sparsification methods that require computationally expensive effective-resistance calculations, DSpar approximates edge importance using only node degree information. For an edge $e=(u,v)$, the sampling probability is defined as

\begin{equation}
p_e \propto \frac{1}{d_u} + \frac{1}{d_v},
\label{eq:dspar}
\end{equation}
where $d_u$ and $d_v$ denote the degrees of nodes $u$ and $v$, respectively. The resulting sparsified graph contains fewer edges while preserving important structural properties of the original graph. Consequently, the number of non-zero entries in the adjacency matrix is reduced, lowering the computational workload of the SpMM operation and enabling more efficient GCN inference.

\section{FPGA-Based GCN Accelerator}\label{sec:proposed}
\subsection{Accelerator Overview}

The proposed accelerator targets efficient Graph Convolutional Network (GCN) inference on FPGA platforms through the combined use of graph sparsification, fixed-point quantization, and approximate arithmetic. Starting from the GCN layer formulation

\begin{equation}
\mathbf{H}^{(l+1)}=\sigma
\left(
\tilde{\mathbf{A}}
\mathbf{H}^{(l)}
\mathbf{W}^{(l)}
\right),
\label{eq:gcn_layer_method}
\end{equation}
the computation is reordered as
\begin{equation}
\mathbf{H}^{(l+1)}=\sigma
\left(
\tilde{\mathbf{A}}
\left(
\mathbf{H}^{(l)}
\mathbf{W}^{(l)}
\right)
\right),
\label{eq:gcn_reordered_method}
\end{equation}

such that dense feature transformation is performed before graph aggregation. Since practical GCNs typically project high-dimensional node features into a lower-dimensional latent space ($F_{out} \ll F_{in}$), this reformulation significantly reduces both memory traffic and computational complexity during the aggregation stage.

\begin{figure}[t]
\centering
\includegraphics[width=\columnwidth]{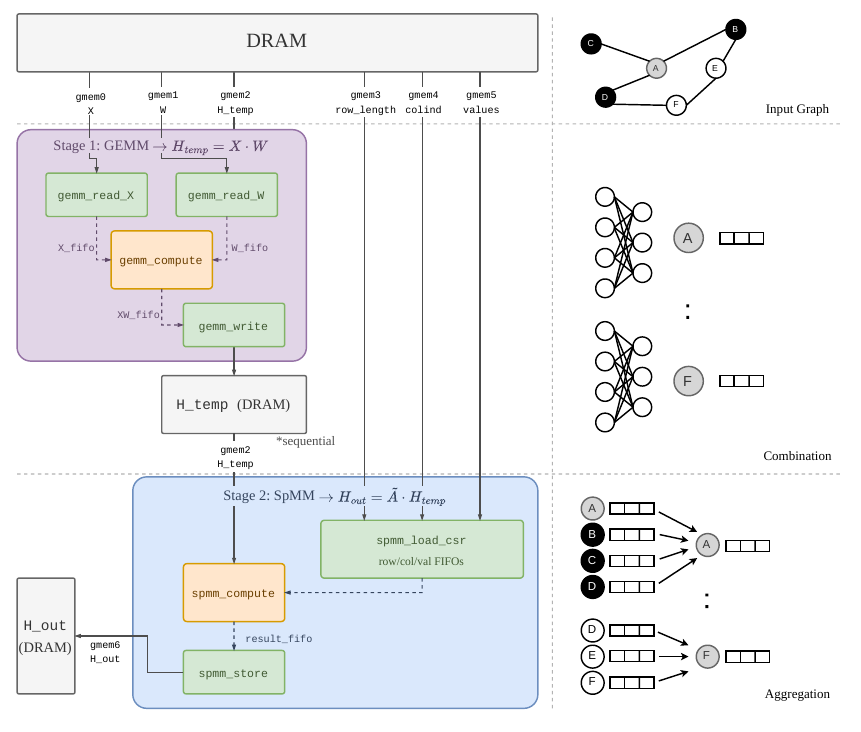}
\caption{System architecture of the proposed FPGA-based GCN accelerator, consisting of a GEMM feature-transformation stage followed by a CSR-based SpMM aggregation stage.}
\label{fig:integrated_gcn}
\end{figure}

Figure~\ref{fig:integrated_gcn} presents the overall architecture of the proposed accelerator. The design consists of two sequential stages connected through an intermediate feature matrix $\mathbf{H}_{temp}$ stored in external memory. Stage~1 performs GEMM to compute transformed node features, while Stage~2 performs SpMM using the graph adjacency structure represented in compressed sparse row (CSR) format. Both stages employ streaming dataflow architectures to overlap communication and computation, while separate AXI interfaces are used for feature matrices, graph metadata, and output buffers to maximize memory-level parallelism.

\subsection{Compute Engines}

The accelerator executes the two stages of (\ref{eq:gcn_reordered_method}) on dedicated streaming engines connected through the intermediate matrix $\mathbf{H}_{temp}$ in external memory. The GEMM engine computes the dense transformation
\begin{equation}
\mathbf{H}_{temp}=\mathbf{X}\mathbf{W},
\label{eq:gemm_method}
\end{equation}
To avoid the memory-bound behavior of a direct evaluation, the matrices are partitioned into tiles of size $M\times M$ and the product is accumulated block-wise as
\begin{equation}
\mathbf{XW}_{ib,jb}
=
\sum_{kb}
\mathbf{X}_{ib,kb}
\mathbf{W}_{kb,jb},
\label{eq:block_gemm}
\end{equation}
enabling on-chip data reuse. Figure~\ref{fig:gemm_figure}
illustrates the resulting GEMM architecture. Within each tile, a $k \rightarrow i \rightarrow j$ execution order maximises reuse while loop unrolling and HLS array partitioning expose parallel multiply-accumulate operations. The engine is split into read, compute, and write stages connected by FIFOs and executed concurrently under the HLS \texttt{DATAFLOW} directive; 512-bit AXI transfers and a transposed, burst-friendly layout of $\mathbf{X}$ further improve effective bandwidth. The proposed design  uses $M=64$ and $U_I=4$, yielding 256 GEMM
MACs per cycle. The SpMM feature loop is unrolled in compile-time.
Larger unrolling factors improve throughput at the cost of memory,
banking, and LUT/DSP~resources.



Following feature transformation, the graph aggregation is performed through a CSR-based SpMM operation.

\begin{equation}
\mathbf{H}_{out}
=
\tilde{\mathbf{A}}
\mathbf{H}_{temp}.
\label{eq:spmm_method}
\end{equation}
Operating on only the non-zero entries of $\tilde{\mathbf{A}}$, its complexity is reduced to
\begin{equation}
\mathcal{O}
\left(
\mathrm{NNZ}
\cdot
F_{out}
\right),
\label{eq:spmm_complexity}
\end{equation}
rather than the $|\mathcal{V}|^2F_{out}$ operations performed for a dense implementation. The adjacency matrix is stored in a modified CSR format that records row lengths explicitly, allowing graph edges to be traversed sequentially and simplifying the dataflow scheduling. As in the GEMM engine, load, compute and store stages run concurrently through FIFOs, with feature-level unrolling across node embeddings; only the feature vector of the neighbor currently being processed is buffered on-chip, keeping the design scalable to larger graphs.







\begin{figure*}[!hbt]
  \centering
  \includegraphics[width=\textwidth]{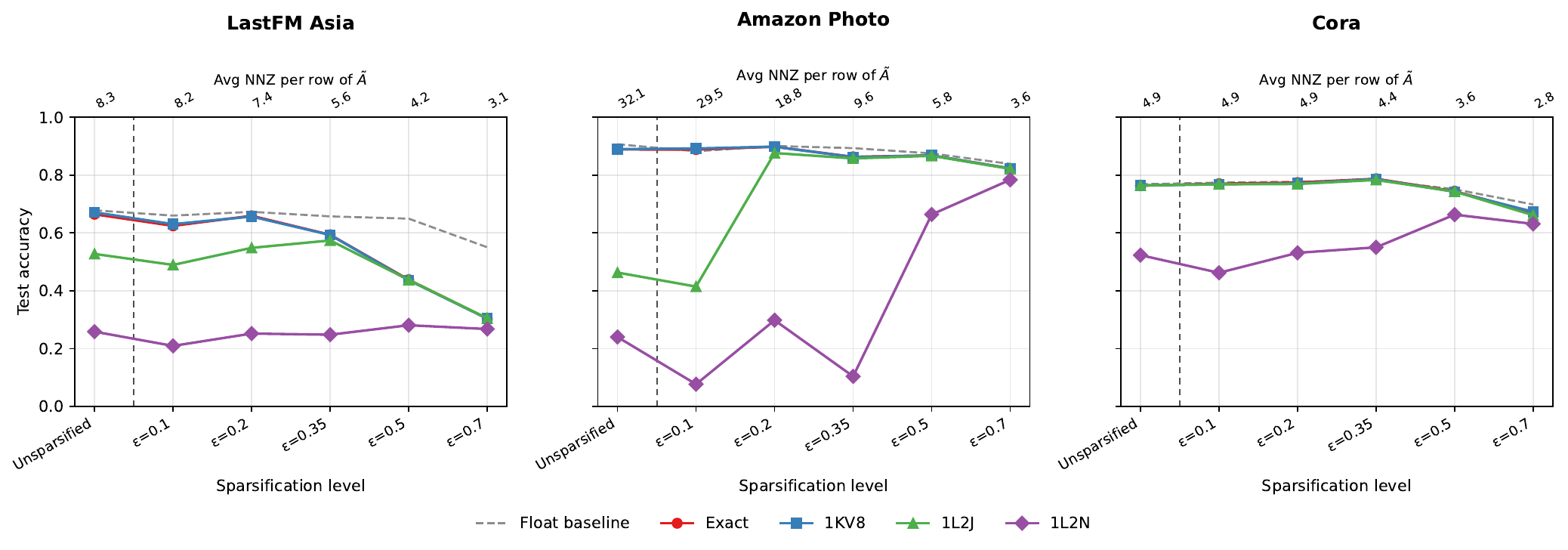}
  \caption{Node-classification test accuracy versus DSpar sparsification level for the exact 8-bit kernel and three EvoApproxLib approximate multipliers (1KV8 near-exact, 1L2J moderate-error, 1L2N high-error) on Cora, Amazon Photo and LastFM Asia. The top axis gives the mean degree (average non-zeros per row of $\tilde{\mathbf{A}}$); the dashed grey line is the FP32 baseline and the vertical divider separates the unsparsified point from the $\epsilon$ sweep. As sparsification lowers per-row accumulation depth, the accuracy lost to approximate multiplication is largely restored, most visibly for 1L2J/1L2N on the denser Amazon Photo graph.}
  \label{fig:accuracy_sweep}
\end{figure*}
\subsection{Graph Sparsification and Approximate Arithmetic}

To further improve efficiency, DSpar is applied as an offline preprocessing step. DSpar removes a subset of graph edges according to degree-aware sampling probabilities while preserving the essential structural characteristics required for message passing. The resulting sparsified adjacency matrix contains fewer non-zero entries, directly reducing the workload of the SpMM stage. Since sparsification is performed prior to deployment, no additional runtime overhead is introduced during inference.

All computations are performed using fixed-point arithmetic to reduce memory footprint and hardware cost. An 8-bit fixed-point implementation was investigated, with wider accumulators employed during reduction operations to prevent overflow. Building on the quantized design, approximate arithmetic is introduced by replacing exact multipliers with approximate multipliers from EvoApproxLib \cite{mrazek2017evoapprox}. Three signed 8-bit multiplier variants -- 1KV8, 1L2J, and 1L2N -- spanning near-exact, moderate-error, and high-error points on the library's power-error Pareto front were evaluated. Each unit is supplied as a synthesizable gate-level C function, which HLS maps into LUT fabric; a lightweight wrapper substitutes it for the \texttt{*} operator at the MAC site, with the active variant selected at compile time. In this work the approximate multiplier is applied only to the SpMM stage of the first GCN layer; the GEMM stage and the entire second layer remain exact.




\section{Experimental Evaluation and Discussion}\label{sec:Exp}
\subsection{Experimental Setup}

The proposed accelerator was evaluated on the AMD Kria KV260 FPGA platform using Vitis HLS at 100 MHz. Experiments were performed on the Cora, LastFM Asia, and Amazon Photo node-classification datasets. These datasets span a wide range of graph densities, from Cora (4.9 average neighbors per node) and LastFM Asia (8.3) to the significantly denser Amazon Photo graph (32.1), providing a suitable testbed for evaluating the interaction between sparsification and approximate computation. The evaluated model is a two-layer GCN, with inference executed layer-by-layer using the proposed HLS kernel. Accuracy is measured against an FP32 baseline, while latency and power are obtained from the FPGA implementation reports.

We evaluate four configurations: an exact 8-bit fixed-point accelerator, an accelerator using approximate multipliers, an accelerator using DSpar graph sparsification, and the combined DSpar plus approximate-multiplier design. Speedup is reported relative to the unsparsified exact 8-bit implementation for each dataset. Latency, speedup, and classification
accuracy are summarized in Table~\ref{tab:summary_results}.

\subsection{Results and Discussion}

Compared with the dense baseline, the optimized 8-bit accelerator
achieves over 1000$\times$ speedup while maintaining baseline
classification accuracy. On the KV260, the proposed CSR-based sparse 8-bit implementation occupies 16.2\% of
LUTs, 8.9\% of FFs, 28.1\% of BRAMs, and 23.2\% of DSPs,
with a total power consumption of 0.668W.

Figure~\ref{fig:accuracy_sweep} shows node-classification accuracy across the full sparsification sweep for all three datasets and the four multiplier configurations. Two key trends emerge. First, the exact and 1KV8 kernels track the FP32 baseline closely until sparsification removes sufficient edges to disrupt message passing: on the sparse Cora graph this occurs beyond $\epsilon = 0.35$, whereas the denser Amazon Photo graph tolerates sparsification up to~$\epsilon = 0.7$.

Second, the impact of approximation scales with graph density. On sparse graphs such as Cora, approximation causes only minor degradation, whereas on Amazon Photo aggressive approximation leads to substantial accuracy loss unless sparsification is applied. As sparsification reduces average node degree, the accuracy of approximate multipliers is progressively restored. LastFM Asia exhibits intermediate behavior between these two extremes. This trend supports the hypothesis that approximation sensitivity is governed by accumulation depth rather than
dataset-specific characteristics.

\begin{table}[t]
\centering
\caption{Latency, speedup and accuracy for the 8-bit quantized GCN kernel with DSpar sparsification only and DSpar \& approximate multiplication combined. Approximate rows use the 1L2J multiplier applied to the SpMM stage of layer~1 only.}
\label{tab:summary_results}
\small
\resizebox{\columnwidth}{!}{%
\begin{tabular}{lcccc}
\toprule
Configuration & Dataset & Latency (ms) & Speedup & Accuracy (\%) \\
\midrule
8-bit Exact            & Cora   & 103.2  & 1.00$\times$ & 76.3 \\
DSpar ($\epsilon=0.35$) & Cora   & 73.9   & 1.40$\times$ & 78.7 \\
DSpar + Approx (1L2J)  & Cora   & 67.8   & 1.52$\times$ & 78.3 \\
\midrule
8-bit Exact            & LastFM & 339.2     & 1.00$\times$ & 66.4 \\
DSpar ($\epsilon=0.35$) & LastFM & 189.7     & 1.79$\times$           & 59.3 \\
DSpar + Approx (1L2J)  & LastFM & 171.6     & 1.98$\times$           & 57.4 \\
\midrule
8-bit Exact            & Photo  & 2216.1 & 1.00$\times$ & 88.9 \\
DSpar ($\epsilon=0.5$)  & Photo  & 247.0  & 8.97$\times$ & 86.8 \\
DSpar + Approx (1L2J)  & Photo  & 224.4  & 9.88$\times$ & 86.6 \\
\bottomrule
\end{tabular}}
\end{table}


\section{Conclusions}\label{sec:Concl}
This paper presented an FPGA-based GCN accelerator that combines graph sparsification, 8-bit quantization, and approximate arithmetic within a unified HLS framework. Experimental results show that graph sparsification and approximate computation are complementary optimization techniques: sparsification reduces aggregation workload while also improving the viability of approximate multipliers by reducing accumulation depth. Evaluated on Cora, LastFM Asia, and Amazon Photo, the proposed design achieves up to 9.88$\times$ speedup while maintaining 86.6\% classification accuracy on Amazon Photo, with total power consumption below 1~W. These results demonstrate that jointly exploiting graph-level and arithmetic-level approximations can enable efficient low-power GCN inference on edge FPGA platforms.

\bibliographystyle{IEEEtran}
\bibliography{refs}

@article{corso2024graph,
  title={Graph neural networks},
  author={Corso, Gabriele and Stark, Hannes and Jegelka, Stefanie and Jaakkola, Tommi and Barzilay, Regina},
  journal={Nature Reviews Methods Primers},
  volume={4},
  number={1},
  pages={17},
  year={2024},
  publisher={Nature Publishing Group UK London}
}

@inproceedings{wu2019simplifying,
  title={Simplifying graph convolutional networks},
  author={Wu, Felix and Souza, Amauri and Zhang, Tianyi and Fifty, Christopher and Yu, Tao and Weinberger, Kilian},
  booktitle={International conference on machine learning},
  pages={6861--6871},
  year={2019},
  organization={Pmlr}
}

@article{zeng2025edge,
  title={Edge graph intelligence: Reciprocally empowering edge networks with graph intelligence},
  author={Zeng, Liekang and Ye, Shengyuan and Chen, Xu and Zhang, Xiaoxi and Ren, Ju and Tang, Jian and Yang, Yang and Shen, Xuemin},
  journal={IEEE Communications Surveys \& Tutorials},
  volume={27},
  number={6},
  pages={3417--3454},
  year={2025},
  publisher={IEEE}
}

@inproceedings{wu2022skeletongcn,
  title={Skeletongcn: a simple yet effective accelerator for gcn training},
  author={Wu, Chen and Tao, Zhuofu and Wang, Kun and He, Lei},
  booktitle={2022 32nd International Conference on Field-Programmable Logic and Applications (FPL)},
  pages={445--451},
  year={2022},
  organization={IEEE}
}

@article{tao2022lw,
  title={{LW-GCN}: A lightweight {FPGA}-based graph convolutional network accelerator},
  author={Tao, Zhuofu and Wu, Chen and Liang, Yuan and Wang, Kun and He, Lei},
  journal={ACM Transactions on Reconfigurable Technology and Systems},
  volume={16},
  number={1},
  pages={1--19},
  year={2022},
  publisher={ACM New York, NY}
}

@inproceedings{zhang2021boostgcn,
  title={{BoostGCN}: A framework for optimizing {GCN} inference on {FPGA}},
  author={Zhang, Bingyi and Kannan, Rajgopal and Prasanna, Viktor},
  booktitle={2021 IEEE 29th Annual International Symposium on Field-Programmable Custom Computing Machines (FCCM)},
  year={2021}
}

@article{romero2021ace,
  title={{ACE-GCN}: A Fast data-driven FPGA accelerator for {GCN} embedding},
  author={Romero Hung, Jos{\'e} and Li, Chao and Wang, Pengyu and Shao, Chuanming and Guo, Jinyang and Wang, Jing and Shi, Guoyong},
  journal={ACM Transactions on Reconfigurable Technology and Systems (TRETS)},
  volume={14},
  number={4},
  pages={1--23},
  year={2021},
  publisher={ACM New York, NY}
}

@inproceedings{nair2023fpga,
  title={Fpga acceleration of gcn in light of the symmetry of graph adjacency matrix},
  author={Nair, Gopikrishnan Raveendran and Suh, Han-Sok and Halappanavar, Mahantesh and Liu, Frank and Seo, Jae-sun and Cao, Yu},
  booktitle={2023 Design, Automation \& Test in Europe Conference \& Exhibition (DATE)},
  pages={1--6},
  year={2023},
  organization={IEEE}
}

@article{liu2023dspar,
  title={{DSpar}: An embarrassingly simple strategy for efficient GNN training and inference via degree-based sparsification},
  author={Liu, Zirui and Zhou, Kaixiong and Jiang, Zhimeng and Li, Li and Chen, Rui and Choi, Soo-Hyun and Hu, Xia},
  journal={Transactions on Machine Learning Research},
  year={2023}
}

@INPROCEEDINGS{mrazek2017evoapprox,
  author={Mrazek, Vojtech and Hrbacek, Radek and Vasicek, Zdenek and Sekanina, Lukas},
  booktitle={Design, Automation \& Test in Europe Conference \& Exhibition (DATE), 2017}, 
  title={EvoApprox8b:  Library of Approximate Adders and Multipliers for Circuit Design and Benchmarking of Approximation Methods}, 
  year={2017},
  volume={},
  number={},
  pages={258-261},
  doi={10.23919/DATE.2017.7926993}}

\end{document}